\documentclass[letterpaper, 10 pt, conference]{ieeeconf}  

\IEEEoverridecommandlockouts                              

\title{\LARGE \bf
Zero123-6D: Zero-shot Novel View Synthesis\\
for RGB Category-level 6D Pose Estimation
}

\author{
    Francesco Di Felice$^{1*}$
    \qquad  
    Alberto Remus$^{1*\dagger}$
    \qquad 
    Stefano Gasperini$^{2}$
    \qquad 
    Benjamin Busam$^{2}$\\
    \qquad 
    Lionel Ott$^{3}$
    \qquad 
    Federico Tombari$^{2,4}$
    \qquad 
    Roland Siegwart$^{3}$
    \qquad 
    Carlo Alberto Avizzano$^{1}$
    \thanks{This project is funded by Leonardo Company S.p.A. under grant No. LDO/CTI/P/0026995/21, July 2\textsuperscript{nd}, 2021.}
    \thanks{$^{*}$ Denotes equal contribution.}
    \thanks{$^{\dagger}$ Corresponding Author (\tt\small alberto.remus95@gmail.com)}
    \thanks{$^{1}$ Department of Excellence in Robotics \& AI, Mechanical Intelligence Institute, Scuola Superiore Sant'Anna, Pisa, Italy ({email: \tt\small firstname.lastname@santannapisa.it}).}
    \thanks{$^{2}$ TUM School of Computation, Information and Technology, Technical University of Munich, Germany.}
    \thanks{$^{3}$ Department of Mechanical and Process Engineering, Autonomous Systems Lab, ETH Zurich, Switzerland.}
    \thanks{$^{4}$ Google Zurich, Switzerland.}
}

\usepackage{graphicx}
\usepackage{amsmath}
\usepackage{amssymb}
\usepackage{mathptmx}
\usepackage{booktabs}
\usepackage{comment}
\usepackage{ulem}
\usepackage{multirow}
\usepackage{color}
\usepackage[table,x11names]{xcolor}
\usepackage[absolute]{textpos}
\definecolor{table_line}{RGB}{222,227,235}

\begin{document}

\begin{textblock}{13}(1.5,0.25)
\centering \noindent\footnotesize © 2024 IEEE.  Personal use of this material is permitted.  Permission from IEEE must be obtained for all other uses, in any current or future media, including reprinting/republishing this material for advertising or promotional purposes, creating new collective works, for resale or redistribution to servers or lists, or reuse of any copyrighted component of this work in other works.
\end{textblock}

\maketitle
\thispagestyle{empty}
\pagestyle{empty}

\begin{abstract}
Estimating the pose of objects through vision is essential to make robotic platforms interact with the environment. Yet, it presents many challenges, often related to the lack of flexibility and generalizability of state-of-the-art solutions. Diffusion models are a cutting-edge neural architecture transforming 2D and 3D computer vision, outlining remarkable performances in zero-shot novel-view synthesis. Such a use case is particularly intriguing for reconstructing 3D objects. However, localizing objects in unstructured environments is rather unexplored. To this end, this work presents \textit{Zero123-6D}, the first work to demonstrate the utility of Diffusion Model-based novel-view-synthesizers in enhancing RGB 6D pose estimation at category-level, by integrating them with feature extraction techniques. Novel View Synthesis allows to obtain a coarse pose that is refined through an online optimization method introduced in this work to deal with intra-category geometric differences.
In such a way, the outlined method shows reduction in data requirements, removal of the necessity of depth information in zero-shot category-level 6D pose estimation task, and increased performance, quantitatively demonstrated through experiments on the CO3D dataset.

\end{abstract}

\section{Introduction}
\label{sec:intro}

The task of 6D pose estimation involves determining the precise 3D position and orientation of objects. It is crucial for numerous applications, including robotics and augmented reality  \cite{gnn_il}. While instance-level approaches have achieved impressive results, their reliance on specific, known objects, typically given by a 3D CAD model, limits their applicability in real-world problems where manual 3D annotations are costly and cumbersome. The assumption of abundant data availability in real-world scenarios is often impractical. Category-level pose estimation has emerged as a promising alternative. It obviates the need for detailed knowledge of individual object instances and allows models to generalize across objects of the same class. Although category-level pose estimation is a well-established field, recent interest has been renewed, aligning it with other successful computer vision tasks like object detection and instance segmentation.
\begin{figure}
    \begin{center}
        \includegraphics[width=0.9\linewidth]{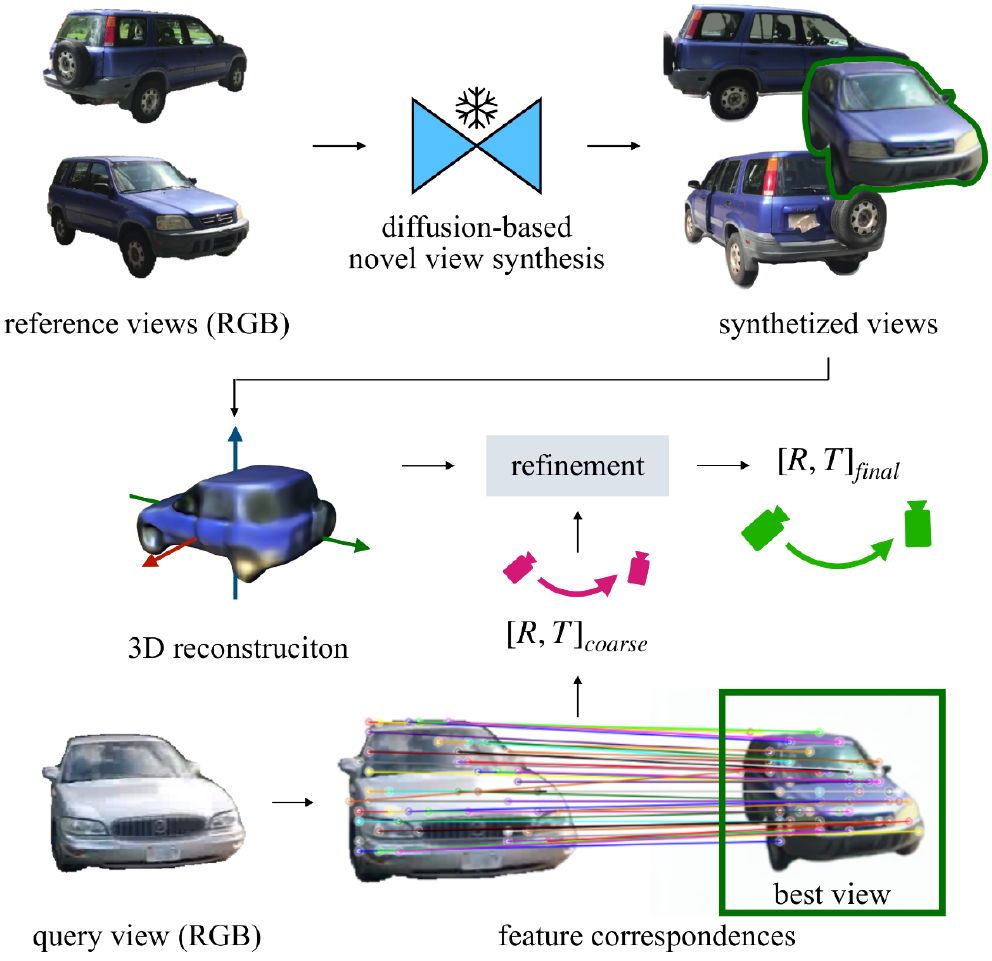}
        \caption{\textbf{Graphical overview of Zero123-6D}. A set of $N$ reference views of instances belonging to a category (top left) is expanded with a novel-view synthesizer (mid). The view that best semantically matches with the query input (bottom left) is selected, while A 3D CAD model is reconstructed from all generated images and their poses. Finally, 2D-3D correspondences are established to refine the best view's estimated pose (bottom right).  
        }
    \vspace{-25pt}
    \end{center} \label{teaser}
  
\end{figure}
Given the data requirements for generalization to unseen objects, recent efforts have focused on simplifying object pose estimation through online optimization, reframing the task as relative pose estimation between a novel query object and a reference. However, existing methods still rely on full scans \cite{sun2022onepose,he2022oneposeplusplus}, CAD models, \cite{ornek2023foundpose,caraffa2023object}, or both \cite{wen2023foundationpose, z6d} for each new instance, which limits their usability in practice.

Some state-of-the-art methods \cite{goodwin2022,goodwin2023you} try to overcome such challenges by leveraging the recent improvements in Foundation Models \cite{caron2021emerging,oquab2023dinov2} to establish semantic correspondences between different instances.
Such approaches are limited in the number and variability of available camera viewpoints, hampering their generalization capabilities in more challenging scenarios. ZSP~\cite{goodwin2022} has demonstrated that an alignment in the feature space between two objects leads to a pose alignment. However, the task of finding the reference view (among an available set) that best aligns with the single-view query image can be challenging if there is not enough viewpoint variability in the reference set.


To address the aforementioned limitations, this document presents Zero123-6D: it exploits pre-trained diffusion models for zero-shot novel-view RGB synthesis of object images in the wild~\cite{zero123, eschernet, zero123++} to expand the available set of reference views. Except for the CAD reconstruction phase (which is performed once per category), none of the models employed in the framework requires any training phase. The overall pipeline just requires a monocular input query of an object instance and a sparse set of posed RGB images of a reference object from the same category. Poses are estimated through feature extraction with a foundation model and online pose optimization based on established feature correspondences.
Zero123-6D reduces previous data requirements and enhances applicability in diverse settings thanks to the pre-trained novel-view synthesizer. To the best of our knowledge, this is the first work that tries to employ a zero-shot diffusion model as a novel-view synthesizer to address the data availability in the context of category-level pose estimation. Even if only RGB images are exploited with no depth information, the pipeline shows increased performance with respect to the existing state of the art.
The main contributions of this work can be summarized as follows:
\begin{itemize}
    
    \item the introduction of diffusion models for zero-shot novel view synthesis to augment the sparse set of reference views available;
    \item a monocular RGB category-level pose estimation method improving upon the current state of the art;
    \item a novel refinement technique to compensate intra-class geometrical variability.
\end{itemize}

The rest of the paper is organized as follows: section \ref{sec:related_work} describes the existent and concurrent related works and open challenges; section \ref{sec:methodology} formalizes the problem and details the structure of the presented architecture and the main novelties introduced; section \ref{sec:experimental_results} collects experimental results with qualitative and quantitative comparisons with state-of-the-art approaches; and section \ref{sec:conclusions} is devoted to conclusions by summarizing the contribution of the work.

\section{Related Work}
\label{sec:related_work}



\subsection{Instance-Level Pose Estimation}
In general, 6D pose estimation requires injecting some source of 3D information to handle the ill-posedness of the task. This can come as 3D Computer-Aided Design (CAD) models encompassing vertices and faces. They represent a dense 3D input that is difficult to provide at inference time.  If a model requires the CAD model of the specific instance at inference time, the task falls into the class of \textit{instance-level} pose estimation, a branch that focuses on accuracy by assuming to have access to this 3D structure at training~\cite{PoseCNN,su2022zebrapose,GDR-Net} and/or inference~\cite{shugurov2022osop} time, which reached outstanding performances in recent years \cite{hodan2018bop,sundermeyer2023bop}. The downside of training one model per object is the lack of generalization in case slightly different samples are presented, which can hinder the construction of autonomous systems in unstructured environments. While self-supervised and ad-hoc reconstruction approaches to the instance problem exist, they require complementary labels~\cite{sock2020introducing,li2023nerf}, data modalities~\cite{ruhkamp2024s} or additional training time~\cite{wang2021occlusion,chen2023texpose}. To this end, this work aims to exploit just the information about the \textit{category} of the considered objects, which is supposed to aggregate common geometrical properties. This gives the name to a more challenging problem named \textit{category-level} pose estimation. 

\subsection{Category-level Pose Estimation}
Category-level pose estimation addresses this challenge by enhancing generalizability across multiple objects within a semantic category despite substantial intra-class variability. NOCS \cite{wang2019normalized} presented a new benchmark for this task that showcased relevant improvements starting from \cite{chen2021fs,lin2021dualposenet,i2c-net} to more recent methods like \cite{lin2022selfdpdn,lin2023vinet,chen2023secondpose}. Nevertheless, this class of methods does not scale well because it requires new training and a new dataset for each class, and data is typically limited~\cite{jung2022housecat6d}.
To overcome such problems, OnePose \cite{sun2022onepose} and OnePose++ \cite{he2022oneposeplusplus} deal with novel objects by acquiring a full RGB scan of the item followed by a structure from the motion pipeline.
The main limitation in this case is the need for a full scan and reconstruction for each new object, limiting its applicability in the case of efficient processing requirements.
POPE~\cite{fan2023pope} represents an alternative approach, which employs off-the-shelf methods \cite{sun2021loftr,oquab2023dinov2} to retrieve the camera pose between different views. Nonetheless, at least one camera-to-object estimate is required to make the method useful for robotic applications. To this end, Zero-Shot-Pose (ZSP)~\cite{goodwin2022} aimed to enhance the generalizability of 6D pose estimation across multiple objects within a given semantic class. 
The method depicts a relative pose estimation setting but assumes depth availability for both query and reference objects.
ZSP uses one RGB-D query image and a set of real-world RGB-D views of the reference object to extract semantic feature correspondences across them with a foundation model. Then, it lifts them to 3D with the available depth maps and computes the alignment. Other than depth availability, the main drawback of ZSP is the viewpoint variability in the reference set: if not enough, it can hinder a proper rotation estimation.
In light of this, the focus is on a generative approach using a diffusion model to synthesize novel views from a very sparse RGB-only reference set.
\begin{figure*} [ht]
    \begin{center}
        \includegraphics[width=1.0\textwidth]{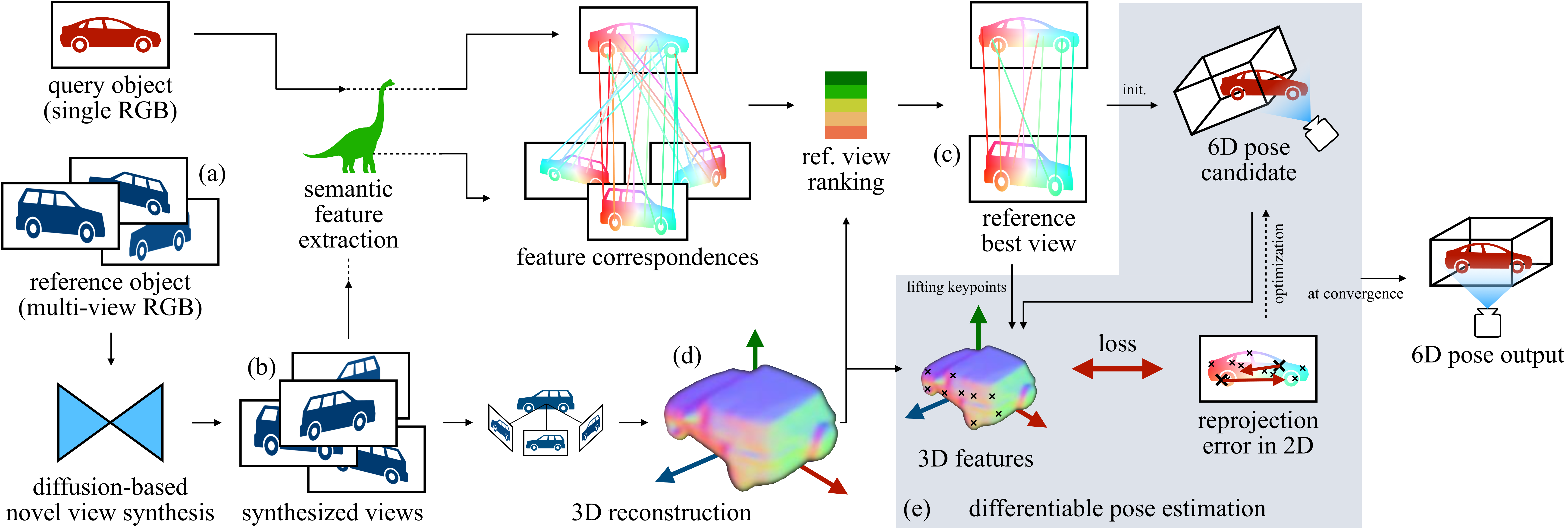}
        \caption{Given a set of $N$ RGB reference views an RGB query image belonging to the same category (a), the goal is to find the 6D pose of the query object. The reference best views are fed to a novel-view-synthesizer diffusion model to generate novel RGB views from coarses poses (b). From that, query and generated views are semantically compared using DINO features, so the reference view that best semantically matches the query is selected, providing a set of 2D correspondences between the 2 views, and a coarse 6D pose (c). At the same time, all the posed generated views of the object are used to reconstruct a 3D mesh using a neural surface reconstructor (d) to obtain the 3D points corresponding to 2D reference matching points. Ultimately, an online optimization process is used between the 2D query points and 3D correspondent reference points to obtain a final refined pose (e).
        } \label{overall_approach}
    \end{center} 
\end{figure*}

\subsection{3D Content Generation and Diffusion Models}
NERF \cite{mildenhall2020nerf} and follow-up works \cite{mueller2022instant} learn an implicit representation of the scene as a radiance field. More recently, 3D Gaussian Splatting \cite{kerbl20233dgaussian} exploits a more 3D grounded representation, where the point cloud of the scene is used to initialize a set of Gaussian that can be rendered to represent high-fidelity scenes. However, the downside of these approaches is that the training can suffer from the availability of a low number of views \cite{eschernet}. 
The rise of 2D generative diffusion models has demonstrated remarkable abilities in the generation of photo-realistic objects and scenes \cite{stable_diffusion,ddpm}. This led to the employment and adaptation of generative diffusion models in 3D content generation domain \cite{magic3d,zhai2024commonscenes}. For dense 3D representations, this can be achieved by leveraging priors encapsulated in 2D diffusion generative models pre-trained on internet-scale datasets. In this context Novel View Synthesis (NVS) domain gained traction thanks to pioneering works like Zero123~\cite{zero123}, which fine-tuned a Latent Diffusion Model (LDM) \cite{stable_diffusion} for zero-shot generation of novel views from just one single view of an object, given a desired camera transformation in spherical coordinates. Building upon these advancements, other diffusion model-based novel view synthesizers emerged aiming to improve the consistency of the generated views for 3D reconstruction from one single view of in-the-wild objects \cite{zero123++,consistent123,one2345,one2345++}. Recently, EscherNet \cite{eschernet}, a multi-view conditioned diffusion model, has shown increased performance in both novel view synthesis and subsequent 3D model reconstruction by leveraging reference-to-reference and target-to-reference consistency layers. Following this direction, in this work, the pre-trained EscherNet is exploited to expand the reference set of views. The model is controlled by means of azimuth, elevation, and radius, which are computed from ground-truth posed images. The views generated by the model are finally used for a NeuS \cite{neus} reconstruction of a 3D CAD model, exploited for the refinement stage.


  

\section{Methodology}\label{sec:methodology}


Our zero-shot pose estimation approach encompasses three steps. First, the reference set is expanded using a zero-shot novel-view synthesizer; all the generated views and their poses are fed inside a 3D neural reconstructor to obtain a 3D CAD model of the object. 
Secondly, the reference object and target object are matched through semantic correspondence by using pre-trained feature extractors \cite{caron2021emerging}; based on these correspondences, the most informative image is selected from the expanded reference sequence. At this point, the pose of the selected view represents the first coarse estimation of the 6D pose, which is used to render its correspondent depth from the reconstructed CAD model. Finally, the coarse pose is refined by leveraging 2D-3D correspondences. Fig. \ref{overall_approach} depicts the schematic overview of the approach.

\subsection{Novel view synthesis} \label{sec:nvs}

The diffusion-model based novel view synthesizer employed in the presented work is EscherNet \cite{eschernet}. Given 1 or more (N) RGB views of an object $\mathbf{I}_{1:N} \in \mathbb{R} ^{H \times W \times 3}$ with their camera poses $\mathbf{P}_{1:N}$ and a set of 1 or more (M) target views $\Tilde{\mathbf{I}}_{1:M}\in \mathbb{R} ^{H \times W \times 3}$ with their desired camera poses $\mathbf{P}_{1:M}$, EscherNet is able to generate multi RGB novel views $\Tilde{\mathbf{I}}_{1:M}$ of the object in the specified poses. The view synthesis is formulated as a conditional generative problem:
\begin{equation}
   \Tilde{\mathbf{I}}_{1:M} \sim p \left( \Tilde{\mathbf{I}}_{1:M} | \mathbf{I}_{1:N}, \mathbf{P}_{1:N}, \mathbf{P}_{1:M} \right)
\end{equation}
The generation of each view $\Tilde{\mathbf{I}}_{1:M}$ depends only on the RGB input views and its relative camera transformation to the input pose of each input view. Once the generation process is completed, RGB images $\Tilde{\mathbf{I}}_{1:M}$ along with their global camera poses $\mathbf{P}_{1:M}$ are used to get the 3D mesh through an implicit-surface reconstructor~\cite{neus}. This model is useful to extract depth information associated with the generated reference views and to refine the pose (\ref{sec:refinement}). Camera poses are encoded by using a spherical coordinate system assuming that the object is centered at the origin of the coordinate system: camera location in the spherical coordinate system is represented by polar, azimuth angle and radius, respectively $\theta,\ \phi,\ r$. Given $(\theta_{1},\ \phi_{1},\ r_{1})$, $(\theta_{2},\ \phi_{2},\ r_{2})$, representing two camera locations, the relative camera transformation between the two cameras is denoted by  $(\Delta \theta =\theta_{2} - \theta_{1},\ \Delta\phi=\phi_{2} - \phi_{1},\ \Delta r= r_{2} - r_{1})$. 




\subsection{Feature extraction and view extraction} \label{sec:feat_extr}

Once the new RGB images are rendered using the diffusion model, it is possible to extract semantic features by relying on various foundation models. Among these DINO \cite{caron2021emerging} has proven to be successfully suitable to extract semantic correspondences between images of different instances. This capability is particularly appealing for category-level tasks, where a semantic relationship between objects of the same class is enforced. As explored in ZSP~\cite{goodwin2022}, 
semantic features can be extracted from image $\mathbf{I_{q}} \in \mathbb{R}^{H \times W \times 3}$ and a reference $\mathbf{I_{r}} \in \mathbb{R}^{H \times W \times 3}$, with the operator $\mathbf{F}: \mathbb{R}^{H \times W \times 3} \rightarrow \mathbb{R}^{h'\times w' \times c}$, with  $h'$,  $w'$, $c$ height, width and channel size of the feature map. These features can be leveraged to establish matches between a query using a soft nearest neighbor operator (SNN):
\begin{equation}
   \text{SNN} \left( \mathbf{I_{q}}, \mathbf{I_{r}} \right)
   := \{ \left( \mathbf{x}_q^k, \mathbf{x}_r^k \right) \mid \mathbf{x}_q^k \leftrightarrow \mathbf{x}_r^k \},
\end{equation}
where $\mathbf{x}_q^k \leftrightarrow \mathbf{x}_r^k, k \in \{1, \ldots, K \}$ are the 2D pixel coordinates of the top $K$ correspondences determined through $\text{SNN}$, in the query and in the reference, respectively. The K correspondences are computed using the extracted query features $\mathbf{F}_{q} := \mathbf{F}( \mathbf{I}_{q} )  \in \mathbb{R}^{h' \times w' \times c} $ and reference features $\mathbf{F}_{r} := \mathbf{F} (\mathbf{I}_r) \in \mathbb{R}^{h' \times w' \times c}$.
The ranking is determined by the cyclical distance $d$ between features calculated through

\begin{equation}
    \alpha =  \Gamma( \mathbf{F}_{r}, \mathbf{F}_{q} [ \mathbf{x}_q] )) \in \mathbb{R}^{2}
\end{equation}
\begin{equation}
    \beta =  \Gamma( \mathbf{F}_{q}, \mathbf{F}_{r} [\alpha] ) \in \mathbb{R}^{2}
\end{equation}
\begin{equation}
    d \left( \mathbf{x}_q, \mathbf{F}_{q}, \mathbf{F}_{r} \right)
    = \left\lVert \mathbf{x}_q - \beta \right\lVert_{2} \in \mathbb{R}
\end{equation}
where $\mathbf{F}_{j}[\mathbf{x}_j]$ denotes the feature $\mathbf{F}_{j} $ correspondent to coordinate $\mathbf{x}_j$ and the function $\Gamma \left( \mathbf{F}_{i}, \mathbf{F}_{j} \left[\mathbf{x}_j \right] \right) =: \bar{\mathbf{x}}_i$ selects the coordinate $\bar{\mathbf{x}}_i$ on the feature map $\mathbf{F}_{i}$ which is closest in feature space to $\mathbf{F}_{j}[\mathbf{x}_j]$. 

We sort $\mathbf{x}_q^k$ in ascending order of values for $d$. These $K$ best correspondences between query view and reference set can then be utilized to select the best RGB image $\Tilde{\mathbf{I}}_r^{\ast}$ in the reference set with
\begin{equation}
    \Tilde{\mathbf{I}}^{\ast} = \operatorname*{argmin}_{i \in 1:N} \sum_{k=1}^K  d \left( \mathbf{x}_q^k, \mathbf{F}_{q}, \mathbf{F}_{i} \right),
\end{equation}
where the cumulative cyclic distance for the top $K$ feature matches is the lowest.
This selection is essential to provide the reference view closest to the query along with its pose (coarse pose) and finally perform an iterative 6D pose estimation refinement.

\subsection{6D pose refinement} \label{sec:refinement}

The main challenge of 6D pose estimation in the presented setting is the lack of depth information for the reference image, which only provides RGB data. 

However, once many views are synthesized by EscherNet, NeuS \cite{neus} (a NeRF-based reconstruction technique) is exploited to produce a 3D mesh using a ray-marching algorithm. From the mesh, it can be extracted the depth $\mathbf{D}_{r}$ aligned with the best RGB image $\Tilde{\mathbf{I}}_i^{\ast}$ by simply rendering the CAD on the corresponding reference pose.

Considering the category-level setting, methods such as E-PnP \cite{EPnP} cannot be used; therefore, a new approach is needed to refine orientation and position transformations.

\subsubsection{3D Lifting and Pose Optimization}
\label{sec:diff_pose_optim}
Given the depth $\mathbf{D_{r}}$ of the reference image $\Tilde{\mathbf{I}}_r^{\ast}$, its intrinsic matrix $\mathbf{K}_{r}$, and extrinsic parameters $\mathbf{R}_{r}, \mathbf{t}_{r}$, each point $\mathbf{x}_{r}$ belonging to the set of 2D reference matches is unprojected into a 3D point $\mathbf{X}_r$ with:

\begin{equation}
    \mathbf{X}_r = \mathcal{P}^{-1} \left( \mathbf{x}_r, D_{r}, \mathbf{K}_{r}, \mathbf{R}_{r}, \mathbf{t}_{r} \right),
\end{equation}

where $\mathcal{P}$ is the projection operator for a pinhole camera.

Once points are lifted, it is possible to use the matches from features to optimize the coarse pose.
Previous methods~\cite{goodwin2022} could use the Umeyama~\cite{umeyama1991least} approach to align lifted 3D-3D correspondences of both query image and reference sequence. However, no depth map is available for the query RGB view; therefore, a method that can leverage 2D-3D correspondences is needed instead. To this end, the idea is to design a differentiable pose optimizer using an iterative online procedure.

Formally, the problem is to optimize for a translation $\mathbf{t}_{q}$ and rotation $\mathbf{R}_{q}$ given each correspondence $\mathbf{X}_r \leftrightarrow \mathbf{x}_q$.
The translation $\mathbf{t}_{q} \in \mathbb{R}^3$ is represented as a vector in Euclidean space. Since the rotation cannot be parameterized in an Euclidean space of dimension three without discontinuities, a more suitable optimization-friendly 6-dimensional representation is used ~\cite{zhou2019continuity}. The continuity of this rotation representation provides superior performance than axis-angle or quaternion representation in the context of numeric optimization while being more compact than the orthogonal $3 \times 3$ matrix with unit determinant ~\cite{zhou2019continuity}. 

We optimize for both rotation and translation $\mathbf{R} $ and $\mathbf{t}$. Given the rotated 3D reference points at each time step $k$ as $\gamma = \left( \mathbf{R} \mathbf{X}_r^k + \mathbf{t} \right)$, we minimize the error between the 2D reprojected reference points $\mathbf{K}_q\gamma$  (using query intrinsic matrix $\mathbf{K}_q$) and the matched 2D points $\mathbf{x}_q^k$ through:
\begin{equation}
    \left( \mathbf{R}_q, \mathbf{t}_q \right)
    = \operatorname*{argmin}_{\mathbf{R}, \mathbf{t}}
    \sum_{k=1}^K \left\lVert \mathbf{K}_q \gamma - \mathbf{x}_q^k  \right\lVert_{2},
    \label{eq:loss_function}
\end{equation}
where $\mathbf{R}_q, \mathbf{t}_q$ are the resulting query rotation and translation respectively.
This iterative refinement process leverages first-order stochastic gradients calculated with Adam~\cite{kingma2014adam} to gradually align points while optimizing for pose parameters.

\section{Experiments}\label{sec:experimental_results}

\begin{table*}[t]
\centering
\caption{Category-level RGB comparison on the CO3D dataset~\cite{goodwin2022} with varying amount of reference views. Zero123-6D is compared with a variant of ZSP~\cite{goodwin2022} that estimates the essential matrix and does not use depth information on the query and reference views.}

\begin{tabular}{ll|ccc|ccc|ccc}
\toprule
 & & \multicolumn{3}{c|}{\textbf{5 reference views}} & \multicolumn{3}{c|}{\textbf{3 reference views}} & \multicolumn{3}{c}{\textbf{1 reference view}} \\ 
Category & Method & med.err ↓ & acc.15 ↑ & acc.30 ↑ & med.err ↓ & acc.15 ↑ & acc.30 ↑ & med.err ↓ & acc.15 ↑ & acc.30 ↑ \\ 
\midrule
\multirow{2}{*}{\textit{bicycle}} & ZSP w/o depth~\cite{goodwin2022} & 51.3 & 4.0 & 22.0 & 60.1 & 7.0 & 25.0 & 98.4 & 3.0 & 10.0 \\ 
 & Zero123-6D [ours] & \textbf{29.2} & \textbf{18.0} & \textbf{50.0} & \textbf{37.1} & \textbf{13.0} & \textbf{40.0} & \textbf{37.1} & \textbf{10.0} & \textbf{40.0} \\ 
\midrule
\multirow{2}{*}{\textit{car}} & ZSP w/o depth~\cite{goodwin2022} & 84.8 & 8.0 & 24.0 & 62.1 & 8.0 & 23.0 & 99.2 & 6.0 & 12.0 \\ 
 & Zero123-6D [ours] & \textbf{9.3} & \textbf{82.0} & \textbf{97.0} & \textbf{12.1} & \textbf{59.0} & \textbf{85.0} & \textbf{14.1} & \textbf{55.0} & \textbf{76.0} \\ 
\midrule
\multirow{2}{*}{\textit{chair}} & ZSP w/o depth~\cite{goodwin2022} & 71.7 & 8.0 & 21.0 & 90.7 & 2.0 & 12.0 & 98.9 & 5.0 & 15.0 \\ 
 & Zero123-6D [ours] & \textbf{28.7} & \textbf{26.0} & \textbf{52.0} & \textbf{31.8} & \textbf{26.0} & \textbf{45.0} & \textbf{43.5} & \textbf{15.0} & \textbf{38.0}  \\ 
\midrule
\multirow{2}{*}{\textit{laptop}} & ZSP w/o depth~\cite{goodwin2022} & 36.0 & 15.0 & 44.0 & 49.1 & 10.0 & 26.0 & 94.3 & 6.0 & 18.0 \\ 
 & Zero123-6D [ours] & \textbf{21.2} & \textbf{38.0} & \textbf{66.0} & \textbf{20.5} & \textbf{34.0} & \textbf{68.0} & \textbf{28.6} & \textbf{28.0} & \textbf{53.0} \\ 
\midrule
\multirow{2}{*}{\textit{motorcycle}} & ZSP w/o depth~\cite{goodwin2022} & 36.5 & 19.0 & 37.0 & 47.3 & 10.0 & 26.0 & 97.1 & 3.0 & 9.0 \\ 
 & Zero123-6D [ours] & \textbf{24.3} & \textbf{23.0} & \textbf{61.0} & \textbf{27.9} & \textbf{15.0} & \textbf{49.0} & \textbf{41.1} & \textbf{13.0} & \textbf{38.0} \\ 
\midrule
\multirow{2}{*}{average} & ZSP w/o depth~\cite{goodwin2022} & 56.0 & 10.8 & 29.6 & 61.9 & 7.4 & 22.4 & 97.6 & 4.6 & 12.8 \\ 
& Zero123-6D [ours] & \textbf{22.6} & \textbf{37.4} & \textbf{65.2} & \textbf{25.8} & \textbf{29.4} & \textbf{57.4} & \textbf{32.9} & \textbf{24.2} & \textbf{49.0}  \\
\bottomrule
\end{tabular}
\label{tab:against_zsp_rgb}
\end{table*}

Quantitative experiments are carried out on the Common Objects in 3D (CO3D) dataset~\cite{reizenstein2021common}, an image collection of 50 object categories spread over 1.5M annotated frames, in real-world conditions. The five chosen categories in this evaluation are   \textit{bicycle}, \textit{car}, \textit{chair}, \textit{laptop}, \textit{motorcycle}.

The objective of the method is to show the possibility to perform category-level pose estimation from a single RGB query image and a set of few RGB references.

For this reason the translation vector computed in Eq.\ref{eq:loss_function} is obtained up-to an unknown scaling factor.
To this end, the focus is on the rotation error (computed as the Geodesic distance on the $\text{SO}(3)$ manifold of rotation matrices). Such an evaluation is present in several RGB-based pose estimation works like POPE \cite{fan2023pope} and NOPE \cite{nguyen2023nope}, which however are limited to instance-level scenarios. Similarly, the proposed evaluation is contained in ZSP in~\cite{goodwin2022}, which represents the baseline for quantitative comparison. Nevertheless, for real-world applications, when the query's depth is available, the scaling factor can be obtained to opportunely rescale the translation vector coming from Eq.\ref{eq:loss_function}.
The assessment includes median rotation error (in degrees) and accuracy. The latter is computed as the percentage of estimates that fall below a certain rotation threshold. Here, 15 and 30 degrees thresholds have been chosen, following ZSP \cite{goodwin2022}. Results are provided both per category and average over all the categories.
Additionally, qualitative results are presented on the Objectron dataset \cite{ahmadyan2021objectron} to highlight the cross-dataset generalization ability of the proposed method to generalize to diverse RGB-only settings. All the experiments have been run on a single NVidia RTX3090 GPU. For each set of reference images, 50 novel views are generated as a trade-off between generation and 3D model reconstruction timing. The pose estimation optimizer runs 1000 iterations with an early stop on the loss computed as in Eq.\ref{eq:loss_function}. This step takes 2-4 seconds to run, depending on the above stopping criteria. The most time-intensive step is the 3D CAD generation, which, however, is carried out only once per category, so in the long run, it does not influence the real-time factor of the application.

\subsection{Quantitative results}

To analyze the quantitative performance of the method, two main analyses have been carried out: the first one is against the depth-free version of ZSP \cite{goodwin2022}, showing the boost provided by the presented approach in reducing the rotation error; the second one is a comparison against the depth-based version of ZSP (which assumes depth availability both for query and reference).

\subsubsection{Comparison against depth-free approach}

Since the presented approach is RGB-only, it is worth starting a quantitative evaluation against methods employing no-depth images, which generally provide a great geometric prior and coherently improved performance. The concurrent work ZSP presented a depth-free version, based on the Essential Matrix calculation \cite{goodwin2022},\cite{goodwin22_supp}. The Essential Matrix embeds the relative pose information between two RGB views from a set of matches. It is typically used in stereo-vision, yet in this case has been effectively employed, by exploiting only the available sparse semantical correspondences.

Table \ref{tab:against_zsp_rgb} presents a full comparison of the work over all the considered CO3D categories, highlighting superior performance on all the metrics, even when just one reference view is available. The key to this success relies on employing zero-shot novel view synthesis, effectively compensating for lacking a dense set of viewpoints. Further, the mesh reconstruction phase allows the estimation of a dense depth map for the reference view. This map allows us to estimate a much more robust set of 2D-3D correspondences and ultimately performs a refined pose estimation.

\subsubsection{Comparison against RGB-D approach}

\begin{table*}[t]
\centering
\caption{Comparison against the Zero Shot Pose \cite{goodwin2022}, augmented to 50 in our setup. BV means the best view selected through semantic matching before pose optimization. Average over all the categories.}

\begin{tabular}{l|ccc|ccc|ccc}
\toprule
\textbf{} & \multicolumn{3}{c|}{\textbf{5 views}} & \multicolumn{3}{c|}{\textbf{3 views}} & \multicolumn{3}{c}{\textbf{1 view}} \\ 
\textbf{Method} & med.err ↓ & acc.15 ↑ & acc.30 ↑ & med.err ↓ & acc.15 ↑ & acc.30 ↑ & med.err ↓ & acc.15 ↑ & acc.30 ↑ \\ 
\midrule
{ZSP BV} & 41.1 & 15.0 & 44.0 & 49.2 & 10.4 & 29.8 & 89.1 & 4.6 & 15.0 \\ 
{Zero123-6D BV} & \textbf{28.0} & \textbf{27.9} & \textbf{57.4} & \textbf{30.5} & \textbf{20.2} & \textbf{52.6} & \textbf{37.1} & \textbf{17.6} & \textbf{44.0}\\

\bottomrule
\end{tabular}
\label{tab:against_zsp_depth_bv}
\end{table*}

\begin{table*}[t]
\centering
\caption{Comparison against the Depth-based ZSP's refined pose \cite{goodwin2022}, augmented to 50 in our setup. Average over all the categories.}

\begin{tabular}{l|ccc|ccc|ccc}
\toprule
\textbf{} & \multicolumn{3}{c|}{\textbf{5 views}} & \multicolumn{3}{c|}{\textbf{3 views}} & \multicolumn{3}{c}{\textbf{1 view}} \\ 
\textbf{Method} & med.err ↓ & acc.15 ↑ & acc.30 ↑ & med.err ↓ & acc.15 ↑ & acc.30 ↑ & med.err ↓ & acc.15 ↑ & acc.30 ↑ \\ 
\midrule
ZSP (RGB-D) & \textbf{20.4} & \textbf{50.2} & \textbf{61.0} & 26.1 & \textbf{43.0} & 55.2 & 93.9 & 22.0 & 31.2 \\
Zero123-6D  & 22.6 & 37.4 & 65.2 & \textbf{25.8} & 29.4 & \textbf{57.4} & \textbf{32.9} & \textbf{24.2} & \textbf{49.0}   \\ 


\bottomrule
\end{tabular}
\label{tab:against_zsp_depth_ref}
\end{table*}

In case depth is available for both query and reference, it boosts the performance of ZSP when a sufficient number of views is present, despite Zero123-6D having a comparable performance. Table \ref{tab:against_zsp_depth_bv} and Table \ref{tab:against_zsp_depth_ref} outline the performance of the compared methods in two modalities. The first one is the best view mode: the result of the matching phase, in which the most similar image is selected based on the highest semantical similarity described in Sec.\ref{sec:methodology}; the pose associated with such a view is the estimated coarse pose.
It is relevant to note that if Table \ref{tab:against_zsp_depth_bv} is compared to the last row of Table \ref{tab:against_zsp_rgb}, even just this best-view selection on synthesized images is capable of surpassing the depth-free version of ZSP by a large margin.

The second mode, analyzed in Table \ref{tab:against_zsp_depth_bv}, encompasses the full method, where ZSP can show better performance only when enough views are available, by exploiting depth information for both query and reference. As it appears from Table \ref{tab:against_zsp_depth_bv}, with the diffusion-based novel-view synthesis, it is possible to substantially improve results over best view selection carried out in ZSP \cite{goodwin2022}, thanks to having more images available for the matching phase. At the same time, this indicates the reduced sim-to-real gap between real and generated images, highlighting how diffusion models can be useful for this task.
On the other hand, as shown in Table \ref{tab:against_zsp_depth_ref}, the Umeyama-RANSAC depth-based refinement method provided by ZSP increases the performance with respect to its correspondent best-view. This method requires an available and employable depth and boosts ZSP's performance; however, it is eroded in few-view reference scenarios, as it is possible to notice a cross-over point with three views. 
Ultimately, Zero123-6D completes the overtake on  ZSP when just a single reference view is available. When reducing from three views to one, Zero123-6D maintains a proper understanding of the 3D structure of the object. At the same time, the depth cues can no longer compensate for the small number of views, as in the previous cases. This can become relevant in fixed camera settings scenarios, where acquiring multiple views of the reference object is difficult.

\subsection{Qualitative results}

\begin{figure}[b]
    \centering
    \includegraphics[width=1.0\linewidth]{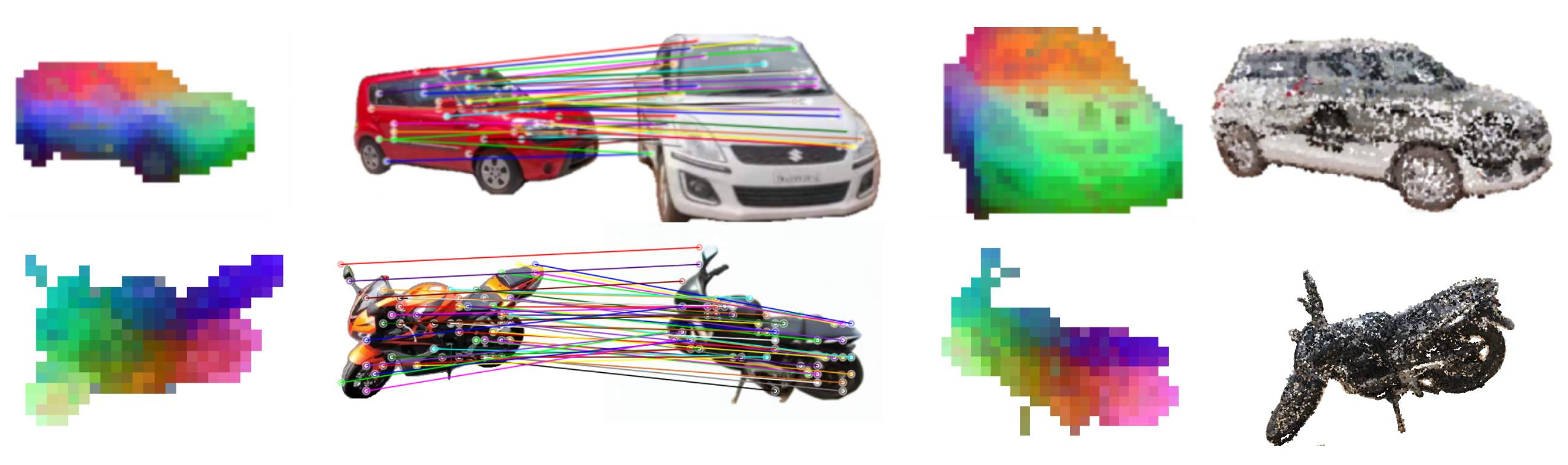}
    \caption{Qualitative results of Zero123-6D on CO3D dataset~\cite{reizenstein2021common} and corresponding feature maps highlighted with PCA at the three channels.}
    \label{fig:qualitative_objectron_feature}
\end{figure}

\begin{figure}[t]
    \centering
    \includegraphics[width=1.0\linewidth]{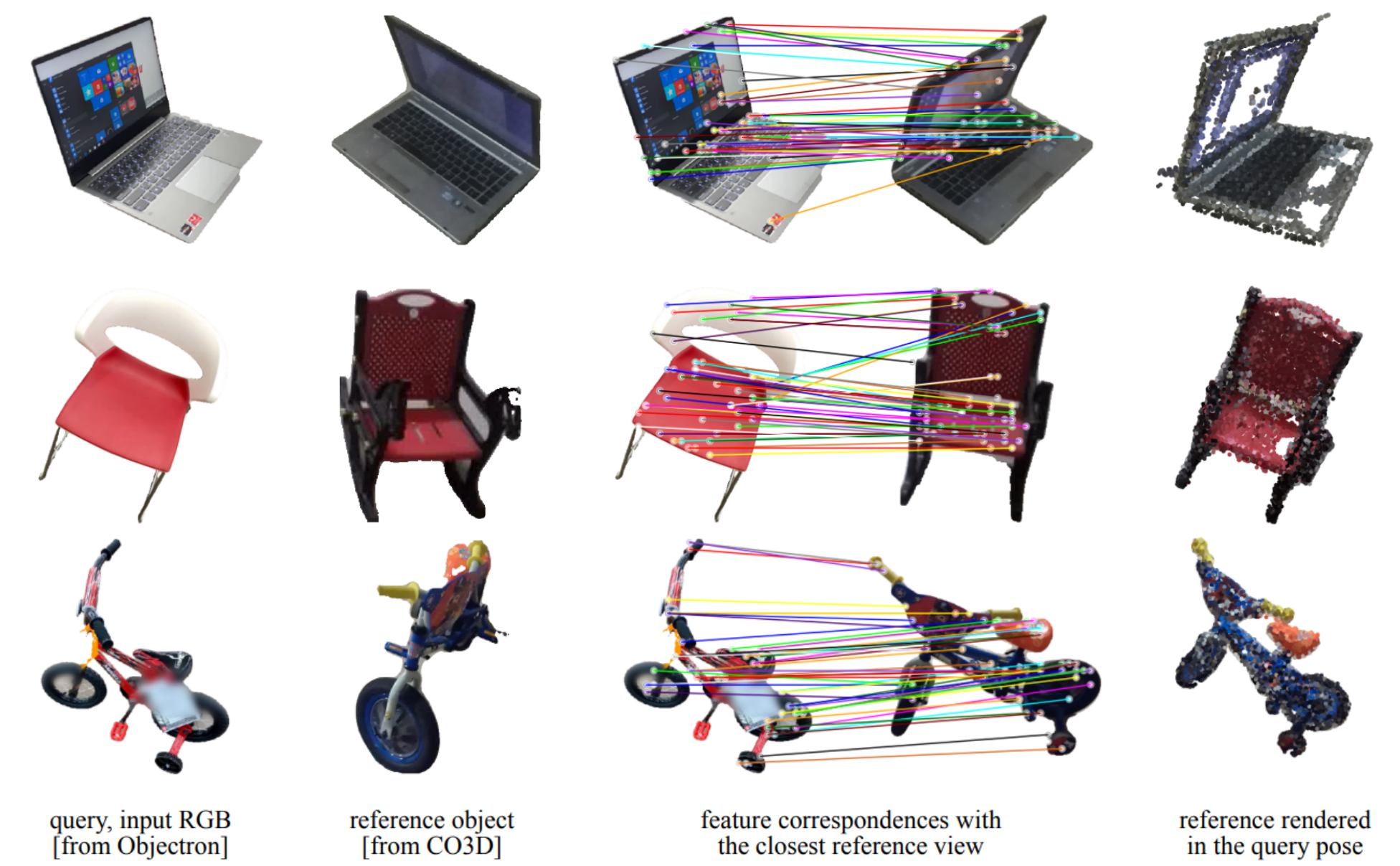}
    \caption{Qualitative results of Zero123-6D on the Objectron dataset~\cite{ahmadyan2021objectron}. From left to right: the query RGB image, a reference example, the closest reference view with matched feature points, and finally, the lifted RGB point cloud of the reference in the estimated 6D pose of the query (only for visualization). The reference objects are from CO3D~\cite{reizenstein2021common}.}
    \label{fig:qualitative_objectron}
    \vspace{-5.1mm}
\end{figure}

This section highlights the qualitative results of the method in accordance with the aforementioned discussion about numerical performance. Fig.\ref{fig:qualitative_objectron_feature} shows the feature map of the instances and the selected best-view. The best-view matching is possible thanks to the semantical information provided by DINO features. Applying Principal Component Analysis (PCA) on the feature map makes it possible to see how similar parts of the objects are depicted with similar colors. This is reflected in the quality of matches, which makes it possible to initialize the presented refiner. As further experiments, qualitative results on the Objectron \cite{objectron2021} dataset are presented. These are particularly interesting because Objectron, unlike CO3D, does not have depth maps associated with RGB images. Hence, in this scenario, methods like ZSP \cite{goodwin2022}, could not be effectively employed without a performance drop, while Zero123-6D's method can be applied unaltered. Results presented in Fig.\ref{fig:qualitative_objectron} depict such cross-dataset generalization, representing a promising direction to employ objects generated through novel-view synthesis for pose estimation in more challenging setups (e.g. cluttered scenarios).

\section{Conclusion and Future Work}\label{sec:conclusions}

This work presented Zero123-6D, a pipeline for category-level pose estimation from a single query image and a set of RGB images belonging to the same class. The proposed online method leverages a limited amount of views, augmented with zero-shot diffusion models, and ultimately improves matching and pose estimates over state-of-the-art solutions. The presented setting can be particularly useful for cases with a limited reference set. Zero123-6D leverages foundation models mainly trained on object-centric datasets \cite{objaverse}; this may limit its efficacy in cluttered scenarios, which can be addressed in future works by extending the proposed method's applicability to other RGB-only settings.

\bibliographystyle{IEEEtran}
\bibliography{biblio}

\begin{thebibliography}{10}
\providecommand{\url}[1]{#1}
\csname url@samestyle\endcsname
\providecommand{\newblock}{\relax}
\providecommand{\bibinfo}[2]{#2}
\providecommand{\BIBentrySTDinterwordspacing}{\spaceskip=0pt\relax}
\providecommand{\BIBentryALTinterwordstretchfactor}{4}
\providecommand{\BIBentryALTinterwordspacing}{\spaceskip=\fontdimen2\font plus
\BIBentryALTinterwordstretchfactor\fontdimen3\font minus
  \fontdimen4\font\relax}
\providecommand{\BIBforeignlanguage}[2]{{%
\expandafter\ifx\csname l@#1\endcsname\relax
\typeout{** WARNING: IEEEtran.bst: No hyphenation pattern has been}%
\typeout{** loaded for the language `#1'. Using the pattern for}%
\typeout{** the default language instead.}%
\else
\language=\csname l@#1\endcsname
\fi
#2}}
\providecommand{\BIBdecl}{\relax}
\BIBdecl

\bibitem{gnn_il}
F.~D. Felice, S.~D'Avella, A.~Remus, P.~Tripicchio, and C.~A. Avizzano,
  ``One-shot imitation learning with graph neural networks for pick-and-place
  manipulation tasks,'' \emph{IEEE Robotics and Automation Letters}, vol.~8,
  no.~9, pp. 5926--5933, 2023.

\bibitem{sun2022onepose}
J.~Sun, Z.~Wang, S.~Zhang, X.~He, H.~Zhao, G.~Zhang, and X.~Zhou, ``Onepose:
  One-shot object pose estimation without cad models,'' in \emph{Proceedings of
  the IEEE/CVF Conference on Computer Vision and Pattern Recognition}, 2022,
  pp. 6825--6834.

\bibitem{he2022oneposeplusplus}
X.~He, J.~Sun, Y.~Wang, D.~Huang, H.~Bao, and X.~Zhou, ``Onepose++:
  Keypoint-free one-shot object pose estimation without {CAD} models,'' in
  \emph{Advances in Neural Information Processing Systems}, 2022.

\bibitem{ornek2023foundpose}
E.~P. {\"O}rnek, Y.~Labb{\'e}, B.~Tekin, L.~Ma, C.~Keskin, C.~Forster, and
  T.~Hodan, ``Foundpose: Unseen object pose estimation with foundation
  features,'' \emph{arXiv preprint arXiv:2311.18809}, 2023.

\bibitem{caraffa2023object}
A.~Caraffa, D.~Boscaini, A.~Hamza, and F.~Poiesi, ``Object 6d pose estimation
  meets zero-shot learning,'' \emph{arXiv preprint arXiv:2312.00947}, 2023.

\bibitem{wen2023foundationpose}
B.~Wen, W.~Yang, J.~Kautz, and S.~Birchfield, ``Foundationpose: Unified 6d pose
  estimation and tracking of novel objects,'' \emph{arXiv preprint
  arXiv:2312.08344}, 2023.

\bibitem{z6d}
P.~Ausserlechner, D.~Haberger, S.~Thalhammer, J.-B. Weibel, and M.~Vincze,
  ``Zs6d: Zero-shot 6d object pose estimation using vision transformers,''
  2023.

\bibitem{goodwin2022}
W.~Goodwin, S.~Vaze, I.~Havoutis, and I.~Posner, ``Zero-shot category-level
  object pose estimation,'' in \emph{Proceedings of the European Conference on
  Computer Vision (ECCV)}, 2022.

\bibitem{goodwin2023you}
\BIBentryALTinterwordspacing
W.~Goodwin, I.~Havoutis, and I.~Posner, ``You only look at one: Category-level
  object representations for pose estimation from a single example,'' in
  \emph{6th Annual Conference on Robot Learning}, 2022. [Online]. Available:
  \url{https://openreview.net/forum?id=lb7B5Rw7tjw}
\BIBentrySTDinterwordspacing

\bibitem{caron2021emerging}
M.~Caron, H.~Touvron, I.~Misra, H.~J{\'e}gou, J.~Mairal, P.~Bojanowski, and
  A.~Joulin, ``Emerging properties in self-supervised vision transformers,'' in
  \emph{Proceedings of the IEEE/CVF international conference on computer
  vision}, 2021, pp. 9650--9660.

\bibitem{oquab2023dinov2}
M.~Oquab, T.~Darcet, T.~Moutakanni, H.~Vo, M.~Szafraniec, V.~Khalidov,
  P.~Fernandez, D.~Haziza, F.~Massa, A.~El-Nouby \emph{et~al.}, ``Dinov2:
  Learning robust visual features without supervision,'' \emph{arXiv preprint
  arXiv:2304.07193}, 2023.

\bibitem{zero123}
R.~Liu, R.~Wu, B.~Van~Hoorick, P.~Tokmakov, S.~Zakharov, and C.~Vondrick,
  ``Zero-1-to-3: Zero-shot one image to 3d object,'' in \emph{Proceedings of
  the IEEE/CVF International Conference on Computer Vision (ICCV)}, October
  2023, pp. 9298--9309.

\bibitem{eschernet}
X.~Kong, S.~Liu, X.~Lyu, M.~Taher, X.~Qi, and A.~J. Davison, ``Eschernet: A
  generative model for scalable view synthesis,'' \emph{arXiv preprint
  arXiv:2402.03908}, 2024.

\bibitem{zero123++}
R.~Shi, H.~Chen, Z.~Zhang, M.~Liu, C.~Xu, X.~Wei, L.~Chen, C.~Zeng, and H.~Su,
  ``Zero123++: a single image to consistent multi-view diffusion base model,''
  2023.

\bibitem{PoseCNN}
Y.~Xiang, T.~Schmidt, V.~Narayanan, and D.~Fox, ``Posecnn: A convolutional
  neural network for 6d object pose estimation in cluttered scenes,'' in
  \emph{Robotics: Science and Systems (RSS)}, 2018.

\bibitem{su2022zebrapose}
Y.~Su, M.~Saleh, T.~Fetzer, J.~Rambach, N.~Navab, B.~Busam, D.~Stricker, and
  F.~Tombari, ``Zebrapose: Coarse to fine surface encoding for 6dof object pose
  estimation,'' in \emph{Proceedings of the IEEE/CVF Conference on Computer
  Vision and Pattern Recognition}, 2022, pp. 6738--6748.

\bibitem{GDR-Net}
G.~Wang, F.~Manhardt, F.~Tombari, and X.~Ji, ``Gdr-net: Geometry-guided direct
  regression network for monocular 6d object pose estimation,'' in \emph{2021
  IEEE/CVF Conference on Computer Vision and Pattern Recognition (CVPR)}, 2021,
  pp. 16\,606--16\,616.

\bibitem{shugurov2022osop}
I.~Shugurov, F.~Li, B.~Busam, and S.~Ilic, ``Osop: A multi-stage one shot
  object pose estimation framework,'' in \emph{Proceedings of the IEEE/CVF
  Conference on Computer Vision and Pattern Recognition}, 2022, pp. 6835--6844.

\bibitem{hodan2018bop}
T.~Hodan, F.~Michel, E.~Brachmann, W.~Kehl, A.~GlentBuch, D.~Kraft, B.~Drost,
  J.~Vidal, S.~Ihrke, X.~Zabulis \emph{et~al.}, ``{BOP}: Benchmark for {6D}
  object pose estimation,'' in \emph{Proceedings of the European Conference on
  Computer Vision}, 2018, pp. 19--34.

\bibitem{sundermeyer2023bop}
M.~Sundermeyer, T.~Hoda{\v{n}}, Y.~Labbe, G.~Wang, E.~Brachmann, B.~Drost,
  C.~Rother, and J.~Matas, ``Bop challenge 2022 on detection, segmentation and
  pose estimation of specific rigid objects,'' in \emph{Proceedings of the
  IEEE/CVF Conference on Computer Vision and Pattern Recognition}, 2023, pp.
  2784--2793.

\bibitem{sock2020introducing}
J.~Sock, G.~Garcia-Hernando, A.~Armagan, and T.-K. Kim, ``Introducing pose
  consistency and warp-alignment for self-supervised 6d object pose estimation
  in color images,'' in \emph{2020 International conference on 3D vision
  (3DV)}.\hskip 1em plus 0.5em minus 0.4em\relax IEEE, 2020, pp. 291--300.

\bibitem{li2023nerf}
F.~Li, S.~R. Vutukur, H.~Yu, I.~Shugurov, B.~Busam, S.~Yang, and S.~Ilic,
  ``Nerf-pose: A first-reconstruct-then-regress approach for weakly-supervised
  6d object pose estimation,'' in \emph{Proceedings of the IEEE/CVF
  International Conference on Computer Vision}, 2023, pp. 2123--2133.

\bibitem{ruhkamp2024s}
P.~Ruhkamp, D.~Gao, N.~Navab, and B.~Busam, ``S2p3: Self-supervised
  polarimetric pose prediction,'' \emph{International Journal of Computer
  Vision}, pp. 1--18, 2024.

\bibitem{wang2021occlusion}
G.~Wang, F.~Manhardt, X.~Liu, X.~Ji, and F.~Tombari, ``Occlusion-aware
  self-supervised monocular 6d object pose estimation,'' \emph{IEEE
  Transactions on Pattern Analysis and Machine Intelligence}, vol.~46, no.~3,
  pp. 1788--1803, 2021.

\bibitem{chen2023texpose}
H.~Chen, F.~Manhardt, N.~Navab, and B.~Busam, ``Texpose: Neural texture
  learning for self-supervised 6d object pose estimation,'' in
  \emph{Proceedings of the IEEE/CVF Conference on Computer Vision and Pattern
  Recognition}, 2023, pp. 4841--4852.

\bibitem{wang2019normalized}
H.~Wang, S.~Sridhar, J.~Huang, J.~Valentin, S.~Song, and L.~J. Guibas,
  ``Normalized object coordinate space for category-level 6d object pose and
  size estimation,'' in \emph{Proceedings of the IEEE/CVF Conference on
  Computer Vision and Pattern Recognition}, 2019, pp. 2642--2651.

\bibitem{chen2021fs}
W.~Chen, X.~Jia, H.~J. Chang, J.~Duan, L.~Shen, and A.~Leonardis, ``{FS-Net}:
  Fast shape-based network for category-level {6D} object pose estimation with
  decoupled rotation mechanism,'' in \emph{Proceedings of the IEEE/CVF
  Conference on Computer Vision and Pattern Recognition}, 2021, pp. 1581--1590.

\bibitem{lin2021dualposenet}
J.~Lin, Z.~Wei, Z.~Li, S.~Xu, K.~Jia, and Y.~Li, ``Dualposenet: Category-level
  6d object pose and size estimation using dual pose network with refined
  learning of pose consistency,'' in \emph{Proceedings of the IEEE/CVF
  International Conference on Computer Vision}, 2021, pp. 3560--3569.

\bibitem{i2c-net}
A.~Remus, S.~D'Avella, F.~D. Felice, P.~Tripicchio, and C.~A. Avizzano,
  ``i2c-net: Using instance-level neural networks for monocular category-level
  6d pose estimation,'' \emph{IEEE Robotics and Automation Letters}, vol.~8,
  no.~3, pp. 1515--1522, 2023.

\bibitem{lin2022selfdpdn}
J.~Lin, Z.~Wei, C.~Ding, and K.~Jia, ``Category-level 6d object pose and size
  estimation using self-supervised deep prior deformation networks,'' in
  \emph{European Conference on Computer Vision}.\hskip 1em plus 0.5em minus
  0.4em\relax Springer, 2022, pp. 19--34.

\bibitem{lin2023vinet}
J.~Lin, Z.~Wei, Y.~Zhang, and K.~Jia, ``Vi-net: Boosting category-level 6d
  object pose estimation via learning decoupled rotations on the spherical
  representations,'' in \emph{Proceedings of the IEEE/CVF International
  Conference on Computer Vision}, 2023, pp. 14\,001--14\,011.

\bibitem{chen2023secondpose}
Y.~Chen, Y.~Di, G.~Zhai, F.~Manhardt, C.~Zhang, R.~Zhang, F.~Tombari, N.~Navab,
  and B.~Busam, ``Secondpose: Se (3)-consistent dual-stream feature fusion for
  category-level pose estimation,'' \emph{arXiv preprint arXiv:2311.11125},
  2023.

\bibitem{jung2022housecat6d}
H.~Jung, S.-C. Wu, P.~Ruhkamp, H.~Schieber, P.~Wang, G.~Rizzoli, H.~Zhao, S.~D.
  Meier, D.~Roth, N.~Navab \emph{et~al.}, ``Housecat6d--a large-scale
  multi-modal category level 6d object pose dataset with household objects in
  realistic scenarios,'' \emph{arXiv preprint arXiv:2212.10428}, 2022.

\bibitem{fan2023pope}
Z.~Fan, P.~Pan, P.~Wang, Y.~Jiang, D.~Xu, H.~Jiang, and Z.~Wang, ``Pope: 6-dof
  promptable pose estimation of any object, in any scene, with one reference,''
  \emph{arXiv preprint arXiv:2305.15727}, 2023.

\bibitem{sun2021loftr}
J.~Sun, Z.~Shen, Y.~Wang, H.~Bao, and X.~Zhou, ``Loftr: Detector-free local
  feature matching with transformers,'' in \emph{Proceedings of the IEEE/CVF
  conference on computer vision and pattern recognition}, 2021, pp. 8922--8931.

\bibitem{mildenhall2020nerf}
B.~Mildenhall, P.~P. Srinivasan, M.~Tancik, J.~T. Barron, R.~Ramamoorthi, and
  R.~Ng, ``Nerf: Representing scenes as neural radiance fields for view
  synthesis,'' in \emph{European Conference on Computer Vision}.\hskip 1em plus
  0.5em minus 0.4em\relax Springer, 2020, pp. 405--421.

\bibitem{mueller2022instant}
T.~M\"uller, A.~Evans, C.~Schied, and A.~Keller, ``Instant neural graphics
  primitives with a multiresolution hash encoding,'' \emph{ACM Trans. Graph.},
  vol.~41, no.~4, pp. 102:1--102:15, Jul. 2022.

\bibitem{kerbl20233dgaussian}
B.~Kerbl, G.~Kopanas, T.~Leimk{\"u}hler, and G.~Drettakis, ``3d gaussian
  splatting for real-time radiance field rendering,'' \emph{ACM Transactions on
  Graphics (ToG)}, vol.~42, no.~4, pp. 1--14, 2023.

\bibitem{stable_diffusion}
R.~Rombach, A.~Blattmann, D.~Lorenz, P.~Esser, and B.~Ommer, ``High-resolution
  image synthesis with latent diffusion models,'' 2021.

\bibitem{ddpm}
J.~Ho, A.~Jain, and P.~Abbeel, ``Denoising diffusion probabilistic models,''
  \emph{arXiv preprint arxiv:2006.11239}, 2020.

\bibitem{magic3d}
C.-H. Lin, J.~Gao, L.~Tang, T.~Takikawa, X.~Zeng, X.~Huang, K.~Kreis,
  S.~Fidler, M.-Y. Liu, and T.-Y. Lin, ``Magic3d: High-resolution text-to-3d
  content creation,'' in \emph{Proceedings of the IEEE/CVF Conference on
  Computer Vision and Pattern Recognition (CVPR)}, June 2023, pp. 300--309.

\bibitem{zhai2024commonscenes}
G.~Zhai, E.~P. {\"O}rnek, S.-C. Wu, Y.~Di, F.~Tombari, N.~Navab, and B.~Busam,
  ``Commonscenes: Generating commonsense 3d indoor scenes with scene graphs,''
  \emph{Advances in Neural Information Processing Systems}, vol.~36, 2024.

\bibitem{consistent123}
H.~Weng, T.~Yang, J.~Wang, Y.~Li, T.~Zhang, C.~L.~P. Chen, and L.~Zhang,
  ``Consistent123: Improve consistency for one image to 3d object synthesis,''
  2023.

\bibitem{one2345}
M.~Liu, C.~Xu, H.~Jin, L.~Chen, Z.~Xu, H.~Su \emph{et~al.}, ``One-2-3-45: Any
  single image to 3d mesh in 45 seconds without per-shape optimization,''
  \emph{arXiv preprint arXiv:2306.16928}, 2023.

\bibitem{one2345++}
M.~Liu, R.~Shi, L.~Chen, Z.~Zhang, C.~Xu, X.~Wei, H.~Chen, C.~Zeng, J.~Gu, and
  H.~Su, ``One-2-3-45++: Fast single image to 3d objects with consistent
  multi-view generation and 3d diffusion,'' \emph{arXiv preprint
  arXiv:2311.07885}, 2023.

\bibitem{neus}
P.~Wang, L.~Liu, Y.~Liu, C.~Theobalt, T.~Komura, and W.~Wang, ``Neus: Learning
  neural implicit surfaces by volume rendering for multi-view reconstruction,''
  \emph{NeurIPS}, 2021.

\bibitem{EPnP}
V.~Lepetit, F.~Moreno-Noguer, and P.~Fua, ``Epnp: An accurate o(n) solution to
  the pnp problem,'' \emph{International Journal of Computer Vision}, vol.~81,
  02 2009.

\bibitem{umeyama1991least}
S.~Umeyama, ``Least-squares estimation of transformation parameters between two
  point patterns,'' \emph{IEEE Transactions on Pattern Analysis \& Machine
  Intelligence}, vol.~13, no.~04, pp. 376--380, 1991.

\bibitem{zhou2019continuity}
Y.~Zhou, C.~Barnes, J.~Lu, J.~Yang, and H.~Li, ``On the continuity of rotation
  representations in neural networks,'' in \emph{Proceedings of the IEEE/CVF
  Conference on Computer Vision and Pattern Recognition}, 2019, pp. 5745--5753.

\bibitem{kingma2014adam}
D.~P. Kingma and J.~Ba, ``Adam: A method for stochastic optimization,''
  \emph{arXiv preprint arXiv:1412.6980}, 2014.

\bibitem{reizenstein2021common}
J.~Reizenstein, R.~Shapovalov, P.~Henzler, L.~Sbordone, P.~Labatut, and
  D.~Novotny, ``Common objects in 3d: Large-scale learning and evaluation of
  real-life 3d category reconstruction,'' in \emph{Proceedings of the IEEE/CVF
  International Conference on Computer Vision}, 2021, pp. 10\,901--10\,911.

\bibitem{nguyen2023nope}
V.~N. Nguyen, T.~Groueix, Y.~Hu, M.~Salzmann, and V.~Lepetit, ``Nope: Novel
  object pose estimation from a single image,'' \emph{arXiv preprint
  arXiv:2303.13612}, 2023.

\bibitem{ahmadyan2021objectron}
A.~Ahmadyan, L.~Zhang, A.~Ablavatski, J.~Wei, and M.~Grundmann, ``Objectron: A
  large scale dataset of object-centric videos in the wild with pose
  annotations,'' in \emph{Proceedings of the IEEE/CVF conference on computer
  vision and pattern recognition}, 2021, pp. 7822--7831.

\bibitem{goodwin22_supp}
W.~Goodwin, S.~Vaze, I.~Havoutis, and I.~Posner, ``Zero-shot category-level
  object pose estimation: Supplementary material,'' in \emph{Proceedings of the
  European Conference on Computer Vision (ECCV)}, 2022.

\bibitem{objectron2021}
A.~Ahmadyan, L.~Zhang, A.~Ablavatski, J.~Wei, and M.~Grundmann, ``Objectron: A
  large scale dataset of object-centric videos in the wild with pose
  annotations,'' \emph{Proceedings of the IEEE Conference on Computer Vision
  and Pattern Recognition}, 2021.

\bibitem{objaverse}
M.~Deitke, D.~Schwenk, J.~Salvador, L.~Weihs, O.~Michel, E.~VanderBilt,
  L.~Schmidt, K.~Ehsani, A.~Kembhavi, and A.~Farhadi, ``Objaverse: A universe
  of annotated 3d objects,'' \emph{arXiv preprint arXiv:2212.08051}, 2022.

\end{thebibliography}

\end{document}